\documentclass{article}

\usepackage[final]{neurips_2022}

\usepackage[utf8]{inputenc} 
\usepackage[T1]{fontenc}    
\usepackage{hyperref}       
\usepackage{url}            
\usepackage{booktabs}       
\usepackage{amsfonts}       
\usepackage{nicefrac}       
\usepackage{microtype}      
\usepackage{xcolor}         
\usepackage{amssymb}
\usepackage{amsmath}
\usepackage{latexsym}
\usepackage{multicol}
\usepackage{booktabs}
\usepackage{multirow}
\usepackage{fancyhdr}
\usepackage{algorithm}
\usepackage{algorithmic}
\usepackage{enumerate}
\usepackage{cancel}
\usepackage{mathrsfs}
\usepackage[compact]{titlesec}
\usepackage{mdwlist}
\usepackage{tikz}
\usepackage{varwidth}
\usepackage[vertfit]{breakurl}
\usepackage{datetime}
\usepackage{pdfpages}
\usepackage{bbm}
\usepackage{tabularx}

\title{Model-based Safe Deep Reinforcement Learning via a Constrained Proximal Policy Optimization Algorithm}

\author{%
  Ashish Kumar Jayant\\
  Department of Computer Science Automation\\
  Indian Institute of Science\\
  Bangalore\\
  \texttt{ashishjayant@iisc.ac.in} \\
  \And
  Shalabh Bhatnagar\\
  Department of Computer Science Automation\\
  Indian Institute of Science\\
  Bangalore\\
  \texttt{shalabh@iisc.ac.in}
}

\begin{document}

\maketitle

\begin{abstract}
 During initial iterations of training in most Reinforcement Learning (RL) algorithms, agents perform a significant number of random exploratory steps. In the real world, this can limit the practicality of these algorithms as it can lead to potentially dangerous behavior. Hence safe exploration is a critical issue in applying RL algorithms in the real world. This problem has been recently well studied under the Constrained Markov Decision Process (CMDP) Framework, where in addition to single-stage rewards, an agent receives single-stage costs or penalties as well depending on the state transitions. The prescribed  cost functions are responsible for mapping undesirable behavior at any given time-step to a scalar value. The goal then is to find a feasible policy that maximizes reward returns while constraining the cost returns to be below a prescribed threshold during training as well as deployment.

We propose an On-policy Model-based Safe Deep RL algorithm in which we learn the transition dynamics of the environment in an online manner as well as find a feasible optimal policy using the Lagrangian Relaxation-based Proximal Policy Optimization. We use an ensemble of neural networks with different initializations to tackle epistemic and aleatoric uncertainty issues faced during environment model learning.  We compare our approach with relevant model-free and model-based approaches in Constrained RL using the  challenging Safe Reinforcement Learning benchmark - the Open AI Safety Gym.  
We demonstrate that our algorithm is more sample efficient and results in lower  cumulative hazard violations as compared to constrained model-free approaches. Further, our approach shows better reward performance than other constrained model-based approaches in the literature. 
\end{abstract}

\section{Introduction}
Deep Reinforcement Learning has provided exceptional results both in the case of
discrete action settings \citep{atari} as well as continuous action domains such as locomotion tasks \citep{haarnoja2018soft}, \citep{schulman2017proximal},
\citep{schulman2018highdimensional}. However, most of the RL algorithms perform significant number of random exploratory steps during
learning as well as deployment which can lead to agents performing undesirable and hazardous
behaviour. This limits the application of RL algorithms in the real world. In 
scenarios like robot navigation \citep{8367110}, 
autonomous driving \citep{kendall2018learning}, healthcare \citep{yu2020reinforcement}, etc., 
where RL has potential applications, unsafe behavior can have hazardous consequences even on human life and property. 

In \cite{surveysafe}, the authors have provided a comprehensive survey of several notions of safety and the associated problem formulations. In our work, we focus on a constraint-based notion of safety. In Constraint-based RL, the goal is to maximize long-term expected reward returns and keep the expected cost-returns below a prescribed threshold. This problem is known as the Safe Exploration problem. In \citep{Achiam2019BenchmarkingSE}, the authors advocate that safety
specifications should be separate from task performance specifications. It also helps in formulating Safe Exploration as a Constrained Optimization Problem and methods used in the optimization
literature \citep{copbook} can be used to solve this problem. Constrained Markov Decision Process (CMDP) \citep{cmdpbook} provides a framework to keep task performance specifications and safety specifications separate from one another. Here in addition to single-stage rewards, state transitions receive single-stage costs as
well. The prescribed cost functions are responsible for mapping undesirable behavior at any given time-step to a non-negative scalar value. 

The existing model-free algorithms used in the Constrained-RL setting suffer from low sample efficiency in terms of environment interactions, i.e., these algorithms require large number of environment interactions (to converge) that in turn would lead to a large  number of hazardous actions due to unsafe exploration. This serves as the motivation for us to use a Model-based approach for Constrained-RL.

\textbf{Our Contributions:} We propose a simple and sample efficient model-based approach for Safe Reinforcement Learning which uses Lagrangian relaxation to solve the constrained RL problem. We highlight the issue that arises due to the use of truncated horizon in Constrained RL and suggest a way to incorporate that in our setting. We demonstrate that our approach is $\sim 3-4$ times more sample efficient than its analogue Model-free Lagrangian relaxation approach and reduces its cumulative hazard violations by $\sim60\%$. Our approach also outperforms model-free Constrained Policy Optimization (CPO) \citep{achiam2017constrained} in terms of constraint satisfaction.  We also compare our approach with model-based approach \citep{LOOP} and we observe that our approach does better in terms of reward performance and has competitive cost performance as well.

\section{Related Work}
Different notions of safety and their mathematical formulations are provided in \cite{surveysafe}.
There are several works on the lines of formulating the Safe RL problem in the setting of CMDP. In some of the early works, an actor-critic algorithm for CMDP under the long-run average cost criterion is proposed in \cite{borkar-cmdp} that is however for the case of full state representations. 
Actor-critic algorithms with linear function approximation have been proposed in \cite{cmdp2, bhatnagaretal} for the long-run average cost setting and in 
\cite{articlecmdpbhatnagar} for the infinite horizon discounted cost scenario. 
The procedure in the aforementioned references 
involved forming a Lagrangian by relaxing the constraints. The algorithms in these papers are based on multi-timescale stochastic approximation with updates of the Lagrange parameter performed on the slow timescale, the policy updates on the medium timescale and the updates of the value function (for a given policy) performed on the fast timescale. 

In more recent work, in \cite{achiam2017constrained}, a trust region based constrained policy optimization (CPO) framework is proposed, which involved approximation of the problem using surrogate functions for both the objective and the constraints and included a projection step on policy parameters that needed backtracking line search, making it complicated and time-consuming. CPO showed near constraint satisfaction in every iteration of the policy updates in standard Mujoco environments modified for safe exploration \citep{achiam2017constrained} 
but didn't yield a constraint satisfying policy in challenging Safe RL benchmark Safety Gym \citep{Achiam2019BenchmarkingSE}. Another work \citep{yu2019convergent} involved using surrogate functions for approximation of both objective and constraint functions. Their procedure involved constructing a sequence of convex optimization problems for which they showed that the sequence of stationary points converges to the stationary point of the original non-convex problem.

In \cite{Achiam2019BenchmarkingSE}, Lagrangian relaxation of the Constrained RL problem is used and combined with PPO \citep{schulman2017proximal} to give a PPO-Lagrangian algorithm and with TRPO \citep{TRPO} to give a TRPO-Lagrangian algorithm. These algorithms were seen to outperform CPO \citep{achiam2017constrained} in terms of constraint satisfaction on several environments in Safety Gym. Also, these algorithms are simpler to implement and tune. Another Lagrangian-based method, see \cite{tessler2018reward}, used a penalized reward function for optimizing their agent and showed convergence to optimal feasible policies using a two-timescale stochastic approximation scheme where the Lagrange multiplier is updated on a slower timescale as compared to the policy parameters as was the case with \cite{borkar-cmdp,articlecmdpbhatnagar,cmdp2}. In \cite{zhang2020order}, the authors proposed a first order constrained policy optimization (FOCOPS) method that involved solving the optimization problem in a non-parametric space and then projecting it back into the  parametric space. Approaches in  \cite{tessler2018reward} and \cite{zhang2020order} performed poorly on Safety Gym. In \cite{stooke2020responsive}, a PID-based approach to damp oscillations in Lagrangian methods is proposed, which is seen to minimize constraint violations. In \cite{crossentropysafe}, the Cross Entropy method \citep{crossentropy} is used for finding a safe policy and convergence using the ODE method is shown, however, empirical results are presented only on a primitive and less challenging environment. In \cite{dalal2018safe}, the authors formulated a state-wise constrained policy optimization problem where at each transition a constraint needs to be satisfied and an analytical method for correcting the unsafe action using a safety layer trained using random exploration was proposed. In \cite{chow2018lyapunov}, the authors proposed constructing Lyapunov functions to guarantee safety of the behaviour policy under a CMDP framework. In \cite{chow2019lyapunov}, the above Lyapunov based method was extended to continuous control but it's  performance in Safety Gym environment  \citep{Achiam2019BenchmarkingSE} was not good in terms of rewards obtained \citep{sikchi2021lyapunov}. In \cite{manifoldliu}, the method involved uses the tangent space of the constraint manifold to learn a safe policy but the approach is  seen to be highly specific to the environment used.

For unconstrained state-of-the-art Model-based RL algorithms in \cite{PILCO,slbo,petshandful,metrpo,svg}, a comparison of the empirical performance on Open AI Gym \citep{gym} is shown in \cite{benchmodelbased}. Also, \cite{LOOP} propose augmenting the planning trajectory with terminal value function to incorporate long-horizon reasoning in model-based methods, since the approaches in
\cite{PILCO,slbo,petshandful,metrpo,svg} 
plan over a fixed and short horizon to avoid aggregation of error. This requires good approximation of both model as well as value function.

There are also several works that use model-based RL to tackle the problem of safety. In \cite{liu2020safe}, a model based approach is proposed to learn the system dynamics and cost model. Then roll-outs from the learned model are used to optimize the policy using a modified cross-entropy based method which involves sampling from a distribution of policies, sorting sample policies based on constraint functions and using them to update the policy distribution. However,  their implementation involves a data collection step using random policy for large number of episodes which itself is risky in real-world scenarios.  In \cite{samba}, model dynamics is learned using PILCO \citep{PILCO} and instead of the discounted cost constraint function, conditional value at risk (CVaR) based constraint function is used \citep{cvar,cvarmdp}. In \cite{imaginingnips}, penalized reward functions are used instead of a separate cost function, then model of the environment is learned and the soft-actor critic algorithm \citep{haarnoja2018soft} is used to optimize the policy. In this approach safe and unsafe states are also needed to be specified upfront.

\section{Background}
\subsection{Constrained Markov Decision Process (CMDP)}
A CMDP is denoted by the tuple $(S,A,R,C,\gamma,\mu)$ where $S$ denotes the state space, $A$ is the
action space, $R: S\times A \times S \rightarrow \mathbb{R}$ is the single-stage reward function, $C: S \times A
\times S \rightarrow \mathbb{R}$ denotes the associated single-stage cost function (we assume a single constraint function for simplicity here), $\gamma$ is the discount factor and $\mu$ signifies the 
initial state distribution. We assume that both $R$ and $C$ are non-negative functions. 

By a policy $\pi=\{\pi_0,\pi_1,\ldots\}$, we mean a decision rule for selecting actions. It is specified as follows: For any $k\geq 0$ and $s\in S$, $\pi_k(s)\in \mathbb{P}(s)$ is the probability distribution $\pi_k(s) \stackrel{\triangle}{=} (\pi_k(s,a),a\in A(s))$ where $\pi_k(s,a)$ is the probability of picking action 
$a$ in state $s$ at instant $k$ under policy $\pi$. In the above, $A(s)$ is the set of feasible actions in state $s$ and so ${\displaystyle A=\cup_{s\in S} A(s)}$. Such a policy is also often referred to as a randomized policy. A stationary policy is a randomized policy as above except with $\pi_k = \pi_l$, $\forall k\not= l$. Thus, a stationary policy selects actions according to a given distribution regardless of the instant when an action is chosen according to the given policy. By an abuse of notation, we denote a stationary policy as $\pi$ itself.

We shall consider here a class of stationary policies $\pi_\theta$ parameterized by a parameter $\theta$. Our  
objective function is defined via the infinite horizon discounted reward criterion where for a given $\pi_\theta$ we have
\begin{equation}
    \label{jr}
J^R(\pi_\theta) = \mathbb{E}\left[\sum_{t=0}^{\infty}
\gamma^t R(s_t,a_t,s_{t+1}) \mid s_0\sim \mu, a_t \sim \pi_\theta, \forall t\right].
\end{equation}
The (cost) constraint function is similarly specified via the following infinite horizon discounted cost:
\begin{equation}
    \label{jc}
J^C(\pi_\theta) = \mathbb{E} \left[\sum_{t=0}^{\infty}
\gamma^t C(s_t,a_t,s_{t+1})\mid s_0\sim \mu, a_t \sim \pi_\theta, \forall t\right].
\end{equation}
Then $J^R(\pi_\theta), J^C(\pi_\theta) \in \mathbb{R}$. Let $d>0$ denote a prescribed threshold below which we want $J^C(\pi_\theta)$ to lie.
The constrained optimization problem then is the following:
\begin{equation}
\label{cop1}
\max_\theta J^R(\pi_\theta) \ \mbox{s.t.}\  J^C (\pi_\theta) \leq d.
\end{equation}
A parameter $\theta$ will be called a feasible point if the cost constraint is satisfied for $\theta$, i.e., $J^C(\pi_\theta) \leq d$.

\subsection{Lagrangian Relaxation based Proximal Policy Optimization}

 The Lagrangian of the constrained optimization problem (\ref{cop1}) can be written as follows:
 \begin{equation}
 \label{lagrangian}
        L(\theta, \lambda) = J^R(\pi_\theta) - \lambda (J^C(\pi_\theta) - d),
    \end{equation}
where $\lambda \in \mathbb{R}^+$ is the Lagrange multiplier and is a positive real number. In terms of the Lagrangian, the goal is to find a tuple $(\theta^*,\lambda^*)$ of the policy and Lagrange parameter such that
\begin{equation}
    \label{lagrangeoptim}
     L(\theta^*,\lambda^*) = \max_\theta \min_\lambda L(\theta,\lambda).
\end{equation}
Solving the max-min problem as above is equivalent
to finding a global optimal saddle point $(\theta^*,\lambda^*)$ 
such that $\forall (\theta,\lambda)$, the following holds:
\begin{equation}
    \label{saddlepointcondition}
 L(\theta^*, \lambda) \geq L(\theta^*, \lambda^*) \geq  L(\theta, \lambda^*).
\end{equation}
We assume that $\theta$ refers to the  parameter of a Deep Neural Network, hence finding such a globally optimal saddle point is computationally hard. So our aim is to find a locally optimal saddle point which satisfies (\ref{saddlepointcondition}) in a local neighbourhood $H_{\epsilon_1,\epsilon_2}$ which is defined as follows:
\begin{equation}
\label{H}
          H_{\epsilon_1,\epsilon_2} \stackrel{\triangle}{=} \{(\theta,\lambda) |\  \|\theta - \theta^*\| \leq \epsilon_1, \|\lambda - \lambda^*\| \leq \epsilon_2\},  \end{equation}
for some $\epsilon_1,\epsilon_2>0$.
Assuming that $L(\theta, \lambda)$ is known for every $(\theta, \lambda)$ tuple,
a gradient search procedure for finding a local $(\theta^*,\lambda^*)$
tuple would be the following:
\begin{eqnarray}
    \theta_{n+1} &=& \theta_n - \eta_1(n)\nabla_{\theta_n} (-L(\theta_n,\lambda_n)), \\
    \label{GDA1}
    &=& \theta_n + \eta_1(n)[\nabla_{\theta_n}J^R(\pi_\theta) -\lambda_n \nabla_{\theta_n}J^C(\pi_\theta)],\\ 
    \lambda_{n+1} &=& [\lambda_n + \eta_2(n)\nabla_{\lambda_n} (-L(\theta_n,\lambda_n))]_+,\\
    \label{GDA2}
    &=& [\lambda_n - \eta_2(n)(J^C(\pi_\theta)-d) ]_+.
\end{eqnarray}

Here $[x]_+$ denotes $\max(0,x)$. This operator ensures that the Lagrange multiplier remains non-negative after each update. In (\ref{GDA1})-(\ref{GDA2}), $\eta_1(n), \eta_2(n)>0$ $\forall n$ are certain prescribed step-size schedules. We assume that the step-sizes $\eta_1(n), \eta_2(n)$, $n\geq 0$ satisfy the regular step-size conditions. Thus, for $i=1,2$,
${\displaystyle 
\sum_k \eta_i(n)=\infty, \sum_k \eta_i^2(n) <\infty}$.
%
Note however that $J^R(\pi_\theta)$ and $J^C(\pi_\theta)$ as specified in (\ref{jr})-(\ref{jc}) are not a priori known quantities and need to be estimated. We discuss this in detail below.

\subsubsection{Estimation}
We run each episode for $T$ time steps in our experiments.
Let $r_{t+1} \equiv R(s_t,a_t,s_{t+1})$ and $c_{t+1} \equiv C(s_t,a_t,s_{t+1})$, respectively, for simplicity. For each sample path we would have both a reward return as well as a cost return. Let $\hat{R}_t$ (resp.~$\hat{C}_t$) be the reward-to-go (resp.~cost-to-go) estimate. We compute $\hat{R}_t$ and $\hat{C}_t$ according to: ${\hat{R}_t = \sum_{k=0}^{T-t-1} \gamma^{k}r_{t+k+1}}$,
${\hat{C}_t = \sum_{k=0}^{T-t-1} \gamma^{k}c_{t+k+1}}$, respectively.
We use a neural network parameterized by $\psi_r$ to estimate reward signal based value function $V_{\psi_{r}}^R$ and a neural network parameterized by $\psi_c$ to estimate a cost signal based value function $V_{\psi_{c}}^C$. We run our simulations on $N$ parallel workers and then sample a mini-batch $\mathcal{M}$ of size $M\leq NT$ \citep{schulman2017proximal}. We use mean-squared loss to estimate value functions on sampled mini-batches as follows:
\begin{eqnarray}
\label{vr}
Loss(\psi_r) = \frac{1}{MT} \sum_{\tau \in \mathcal{M}} \sum_{t=0}^{T} (V_{\psi_{r}}^R(s_t) - \hat{R}_t)^2,\\
\label{vc}
Loss(\psi_c) = \frac{1}{MT} \sum_{\tau \in \mathcal{M}} \sum_{t=0}^{T} (V_{\psi_{c}}^C(s_t) - \hat{C}_t)^2.
\end{eqnarray}
Let $A^R_t$, $t\geq 0$, and
$A^C_t$, $t\geq 0$, respectively, denote the advantage estimates w.r.t reward and cost value functions on the sample path. We compute them using Generalized Advantage Estimation \citep{schulman2018highdimensional} to balance the bias and variance tradeoff of advantage estimates. We can have a range of advantage estimates as under.
\begin{eqnarray}
\label{lowk}
A^1_t = r_{t+1} + \gamma V(s_{t+1}) - V(s_t),\\
A^2_t = r_{t+1} + \gamma r_{t+2} + \gamma^2V(s_{t+2}) - V(s_t),\\
\label{highk}
A^k_t = r_{t+1} + \dots  + \gamma^{k-1}r_{t+k}+ \gamma^kV(s_{t+k}) - V(s_t),
\end{eqnarray}
for $k>2$. The advantage estimate in (\ref{lowk}) will have high bias but low variance while estimates in (\ref{highk}) with a higher value of $k$ generally have high variance but low bias. Let $\delta^R_t = r_{t+1} + V^R_{\psi_r}(s_{t+1}) - V^R_{\psi_r}(s_t)$ and $\delta^C_t = c_{t+1} + V^C_{\psi_c}(s_{t+1}) - V^C_{\psi_c}(s_t)$, respectively, denote the reward and cost temporal differences. Let $\bar{\lambda}$ be a parameter which adjusts the bias-variance tradeoff. Generalized Advantage Estimates \citep{schulman2018highdimensional} for $A^R_t$, $A^C_t$ are then given by
\begin{eqnarray}
\label{AR}
A^R_t = \sum_{l=0}^k (\gamma \bar{\lambda})^l \delta^R_{t+l},\\
\label{AC}
A^C_t = \sum_{l=0}^k (\gamma \bar{\lambda})^l \delta^C_{t+l},
\end{eqnarray}
respectively. Now we use PPO clipped objectives \citep{schulman2017proximal} for estimation of $J^R(\pi_\theta), J^C(\pi_\theta)$ as follows:
\begin{eqnarray}
\label{ppo_jr}
    J^R(\pi_\theta) =  \mathbb{E}_{t}[ \min(r_t(\theta)A^R_{t}, \mbox{clip}(r_t(\theta), 1-\epsilon,1+\epsilon)A^R_{t})],
\end{eqnarray}
\begin{eqnarray}
\label{ppo_jc}
    J^C(\pi_\theta) =  \mathbb{E}_{t}[ \min(r_t(\theta)A^C_{t}, \mbox{clip}(r_t(\theta), 1-\epsilon,1+\epsilon)A^C_{t})],
\end{eqnarray}
where $r_t(\theta) = \frac{\pi_\theta(a_t|s_t)}{\pi_{\theta_{old}}(a_t|s_t)}$ is the ratio of the probability of selecting action $a_t$ in state $s_t$ under parameter $\theta$ as opposed to $\theta_{old}$. Further, $A^R_t$ and $A^C_t$ are the estimated advantages based on the reward and cost returns, respectively, by time $t$ (see above) and $\epsilon$ is the clip-ratio which clips $r_t(\theta)$ to $(1-\epsilon)$ if it is less than $(1-\epsilon)$ and clips to $(1+\epsilon)$ if it is greater than $(1+\epsilon)$. This algorithm restricts the policy parameters to not change significantly between two iterations which helps in avoiding divergence. This approach is referred to as PPO-Lagrangian \citep{Achiam2019BenchmarkingSE}.

\subsection{Model-based Constrained RL}

We  formulate a Constrained RL problem (\ref{copm}) using a model-based framework as follows:
\begin{eqnarray}
\label{copm}
\max_{\pi_\theta \in \Pi_\theta} J_m^R(\pi_\theta) \ 
\mbox{ s.t. }\  J_m^{C}(\pi_\theta) \leq d, \mbox{ where }
\end{eqnarray}
\begin{eqnarray}
\label{jmr}
J_m^R(\pi_\theta) &=& \mathbb{E}\left[\sum_{t=0}^{\infty}
\gamma^t R(s_t,a_t,s_{t+1}) \mid s_0\sim \mu,\ s_{t+1} \sim P_\alpha(.|s_t,a_t),\ a_t \sim \pi_\theta, \forall t\right],\\
\label{jmc}
J_m^{C} (\pi_\theta) &=& \mathbb{E}\left[\sum_{t=0}^{\infty}
\gamma^t C(s_t,a_t,s_{t+1}) \mid s_0\sim \mu,\ s_{t+1} \sim P_\alpha(.|s_t,a_t),a_t \sim \pi_\theta, \forall t\right].
\end{eqnarray}
In the above, $P_\alpha(.|s_t,a_t)$ is an  $\alpha$-parameterized environment model, $d_i$ is a  human prescribed safety threshold for the $i$th constraint and  $\Pi_\theta$ is the set of all $\theta-$parameterized stationary polices. Note that we assume the initial state $s_0$ is sampled from the true initial state distribution $\mu$ and then $s_{t+1} \sim P_\alpha(.|s_t,a_t), \ \forall t>0$.
We would use approximation of environment $P_\alpha$ to create `imaginary' roll-outs to estimate the reward and cost returns required for policy optimization algorithms.

\section{Challenges in Environment Model Learning}
In this section we discuss the challenges that commonly arise due to model learning in RL. We further highlight the challenge that arises from using environment model approximation in Safe RL settings. 
\begin{enumerate}
    \item \textbf{Handling aleatoric and epistemic uncertainties} : \textit{Aleatoric Uncertainty}  refers to the notion of natural randomness in the system which leads to variability in outcomes of an experiment. This uncertainty is irreducible because it is a natural property of the system. Hence in such cases, giving measure of uncertainty in model's prediction is a good practice. In \cite{uncertaintyprediction,petshandful} uncertainty-aware neural networks are used which give an idea about uncertainty in prediction as well. They learn Gaussian distribution parameterized by neural networks. \textit{Epistemic Uncertainty}  refers to the notion of lack of sufficient knowledge in the model as a result of which the model does not generalize. In \cite{uncertaintyprediction}, an ensemble of uncertainty aware neural networks with different initializations is proposed to reduce epistemic uncertainty for fixed data.
    
For the learning environment model, we also use an ensemble of $n$ neural networks with random initialization. Each neural network's output parameterizes a multivariate normal distribution with diagonal covariance matrix.  
Suppose the $i$th neural network in the ensemble is parameterized by $\alpha_i$ and the mean and standard deviation outputs are $\mu_{\alpha_i}$ and $\sigma_{\alpha_i}$ respectively.
Recall now that if a random vector $X \in \mathbb{R}^{d}$ is distributed according to the  multivariate normal distribution parameterized by $(\mu,\Sigma_{d\times d})$ where $\mu \in \mathbb{R}^{d}$ is the mean vector and $\Sigma_{d\times d}$ is a $d\times d$ covariance matrix, then the probability density of $X$ is defined as, $P(x) = \frac{1}{(2\pi)^{d/2}} |\Sigma|^{-1/2} exp ( - \frac{1}{2} (x-\mu)'\Sigma^{-1}(x-\mu)), x\in \mathbb{R}^d$.  

As a choice of loss function we use the negative log-likelihood loss for minimization (i.e., minimizing negative log of $P(X)$). For the $i$th neural network parameterized by $\alpha_i$, the loss function $L(\alpha_i)$ is given as follows: 
\begin{equation}
\label{loss_ensemble}
L(\alpha_i) = \sum_{t=1}^{M}[\mu_{\alpha_i}(s_t,a_t) - s_{t+1}]^{T}\Sigma_{\alpha_i}^{-1}(s_t,a_t)[\mu_{\alpha_i}(s_t,a_t) - s_{t+1}] + \log |\Sigma_{\alpha_i}(s_t,a_t)|,
\end{equation}
where $\mu_{\alpha_i}(s_t,a_t)$ is the mean vector output of the $i$th neural network and $\Sigma_{\alpha_i}(s_t,a_t)$ is the covariance matrix which is assumed to be a diagonal matrix. Note that, using $n$ neural networks with random initialization tends to have a regularization effect. Intuitively, it introduces diversity in learned models to deal with more possible trajectories. Also, it has been shown empirically that with an increase in the number
of models, the performance tends to improve (See section 6.3 and Figure 4 of \cite{metrpo}), but increase in number of models also leads to increase in space complexity.
    
    \item \textbf{Aggregation of Error} : In model-based RL, as we move forward along the horizon, the error due to approximation starts aggregating and predictions from the approximated model tend to diverge significantly from the true model. In order to tackle this problem, most of the model-based RL approaches \citep{PILCO,metrpo,MBPO} use shorter (or truncated) horizon during the policy optimization phase and achieve similar performance as Model-Free RL approaches. We use truncated horizon in our approach.  
    \item \textbf{Implication of using truncated horizon in Constrained RL} : When we use truncated horizon in Constrained RL, it leads to underestimation of cost returns (\ref{jmc}) under the current policy and we use the prescribed constraints to threshold the returns obtained. This can lead to constraint violations in the  real-environment where the cost objective is based on the infinite horizon cost return. We propose a  hyperparameter-based approach to deal with this problem in the next section.
\end{enumerate}

\section{Model-based PPO Lagrangian}
We propose a model-based algorithm obtained  from relaxing the Lagrangian (see Algorithm \ref{alg:ppo_mb}) which alleviates the problem with obtaining a large number of samples in model-free Lagrangian based approaches and as a consequence also decreases the cumulative hazard violations. It is difficult to evaluate the policy without interacting with the real environment accurately.
\begin{figure}[H]
\centering
\begin{algorithm}[H]
\begin{algorithmic}[1]
 \STATE {\bfseries Input:} Initialize actor neural net parameter $\theta_0$, critic parameters $\psi_{r0},\psi_{c0}$, ensemble models $[P_{\alpha_{i}}]^{n}_{i=1}$, Lagrange parameter $\lambda_0\geq 0$, cost threshold = $d$, Environment Horizon = $T$, Model Horizon = $H$
 \FOR{$i =1, \dots, N$\ training\ epochs}
  	
	\STATE Collect data tuples $(s_t,a_t,s_{t+1})$, for the $i$th ensemble 
	using policy $\pi_{\theta_{i}}$, $i=1,\ldots,n$, in the environment for $T$ time steps over multiple ($|E|$) episodes
	\STATE Train $[P_{\alpha_{i}}]^{n}_{i=1}$ by minimizing (\ref{loss_ensemble}) w.r.t $\alpha_i,\ \forall i=1\ to\ n$
 	\WHILE{Performance ratio $> PR_{threshold}$}
 	    \STATE $s_0 \sim \mu$ 
 	    \newline \algorithmiccomment{\textit{Note: For first pass we use a mix of real data and imaginary roll-out data. (See Appendix A)}}
 	    \STATE Collect data roll-outs as $a_t \sim \pi_{\theta_i}(\cdot|s_t),\ s_t \sim P_{\alpha_{q}}(\cdot|s_t,a_t)$
 	    (At each time step '$q$' is randomly selected from $1,\ldots,n$) for $H$ time steps $(H<T)$
     	\STATE  Compute $J_{sample}^C(\pi_{\theta_t}) = \frac{1}{|E|} \sum_{p=1}^{H} \gamma^p C(s_t,a_t)$ where $|E|$ is the number of episodes
     	\STATE Compute advantage, cost-advantage using (\ref{AR}) and (\ref{AC}) respectively
     	\STATE Update $\lambda$ by replacing $J^C(\pi_{\theta})$ with  $J_{sample}^C(\pi_{\theta_i})$ and using an  appropriate value of $\beta$ in (\ref{GDA2_mod}).
     	\newline \algorithmiccomment{\textit{Multiple gradient updates for actor and critic}}
     	\FOR{$k=1,\ldots,K$}
     	\STATE Compute $J^R({\pi_{\theta_k}}),J^C({\pi_{\theta_k}})$ as in (\ref{ppo_jr}) and (\ref{ppo_jc}) respectively.
     	\STATE Update parameters $\theta_k$ using (\ref{GDA1})
     	\STATE Update critic parameters $\psi_r$ and $\psi_c$ minimizing (\ref{vr}) and (\ref{vc}) respectively.
     	\ENDFOR
     	\STATE Compute Performance Ratio ($PR$) using (\ref{pr})
     \ENDWHILE
 \ENDFOR
\end{algorithmic}
\caption{Our Approach: Model-based PPO-Lagrangian}
\label{alg:ppo_mb}
\end{algorithm}
\end{figure}
  For this we compute the Performance Ratio ($PR$) metric using ensemble models (see \cite{metrpo}) that is defined as follows:
 \begin{equation}
            PR = \frac{1}{n} \sum_{i=1}^{n} \mathbbm{1}( \zeta^R(\alpha_i,\theta_t)>\zeta^R(\alpha_i,\theta_{t-1})),
        \label{pr}
        \end{equation}
where $\zeta^R(\alpha_i,\theta_t) = \sum_{t=0}^{T}
\gamma^t R(s_t,a_t,s_{t+1}),\ s_0 \sim \mu\ $ and $ \forall t\geq 0: s_{t+1} \sim P_{\alpha_{i}}(\cdot|s_t,a_t), a_t \sim \pi_{\theta_{t}}(\cdot|s_t)$. This measures the ratio of the number of models in which policy is improved to the total number of models in ensemble $(n)$. If $PR>PR_{threshold}$, we continue training using the same model, if not then we break and re-train our environment model on data collected from the new update policy. 

Another challenge that we encounter is the  underestimation of $J^C(\pi_\theta)$ resulting from using a truncated horizon (of length $H$ in step 7 of Algorithm \ref{alg:ppo_mb}) to reduce the aggregation of error.
\begin{figure}[H]
    \centering
    \includegraphics[width=1.0\linewidth]{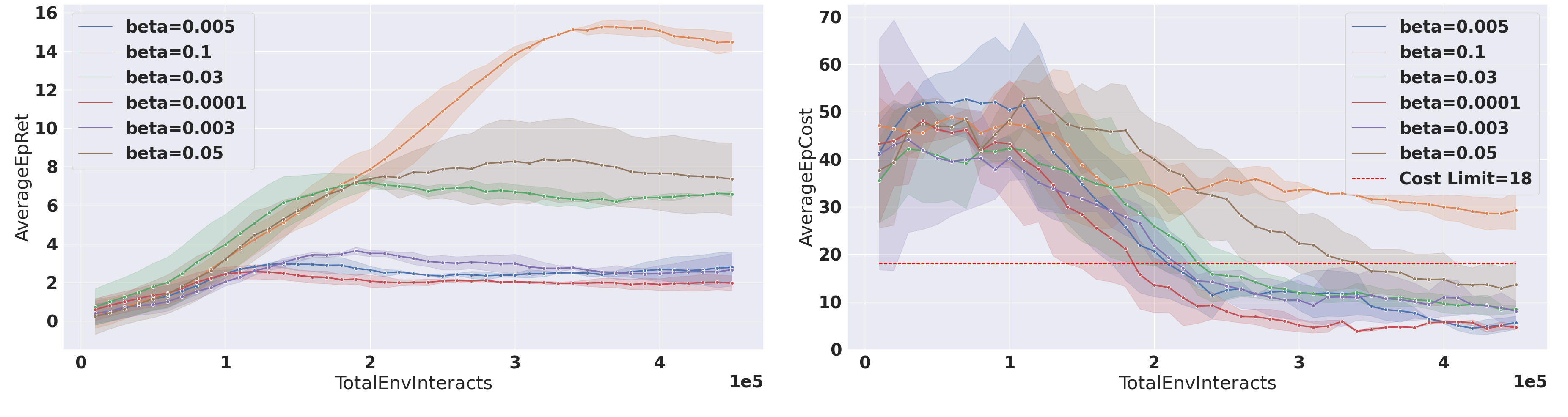}
    \caption{Effect of beta parameter $(\beta) $ on expected cost returns (left) and expected reward returns (right) in PointGoal environment. (Here $\beta=0.1$ corrresponds to $\frac{H}{T}$)}.
    \label{fig:beta}
\end{figure}
This can potentially lead to constraint violating policies. So we need to make the safety threshold $(d)$ stricter. It might seem that changing this threshold $(d)$ proportional to the truncated horizon, i.e., setting $d := d*\frac{H}{T}$ might work better where $H$ is the truncated horizon and $T$ is the original environment horizon, but we found that this leads to constraint violations as well because we are learning policy using data from the  approximated environment model that would bring in errors. Hence the cost estimate of the policy (Step 8 in Algorithm \ref{alg:ppo_mb}) is error-prone as well. To tackle this issue, we make safety threshold stricter using a hyperparameter $0\leq \beta < 1$ by modifying the Lagrange multiplier update as follows:
\begin{eqnarray}
\label{GDA2_mod}
     \lambda_n = [\lambda_n - \eta_2(n)(J^C(\pi_\theta)-d*\beta) ]_+.
\end{eqnarray}
The variation of expected cost returns and reward returns with respect to $\beta$ is shown in Figure \ref{fig:beta} on Safety gym \textit{PointGoal} \citep{Achiam2019BenchmarkingSE} environment. We can observe that as we reduce $\beta$, the expected cost returns reduce because cost limit becomes stricter but choosing too small a $\beta$ leads to low reward returns as well. 

\section{Experimental Details and Results}
We test our approach on Safety Gym environments \footnote{\url{https://github.com/openai/safety-gym}} - \textit{PointGoal} and \textit{CarGoal} with modified state representations as used in other model-based Safe RL approaches \citep{liu2020safe, LOOP} that are more helpful in model-learning. We increase the difficulty of \textit{PointGoal} and \textit{CarGoal} environments by increasing the number of hazards from 10 to 15. In both environments, the aim of robots is to reach the goal position and have as few collisions with hazards as possible. We compare our approach (MBPPO-Lagrangian) with Unconstrained PPO \citep{schulman2017proximal}, model-free Safe RL approaches including Constrained Policy Optimization \citep{achiam2017constrained}, PPO-Lagrangian \citep{Achiam2019BenchmarkingSE} and the model-based approach -- safe-LOOP from \cite{LOOP}. 

\begin{figure}[H]
    \centering
    \includegraphics[width=0.98\linewidth]{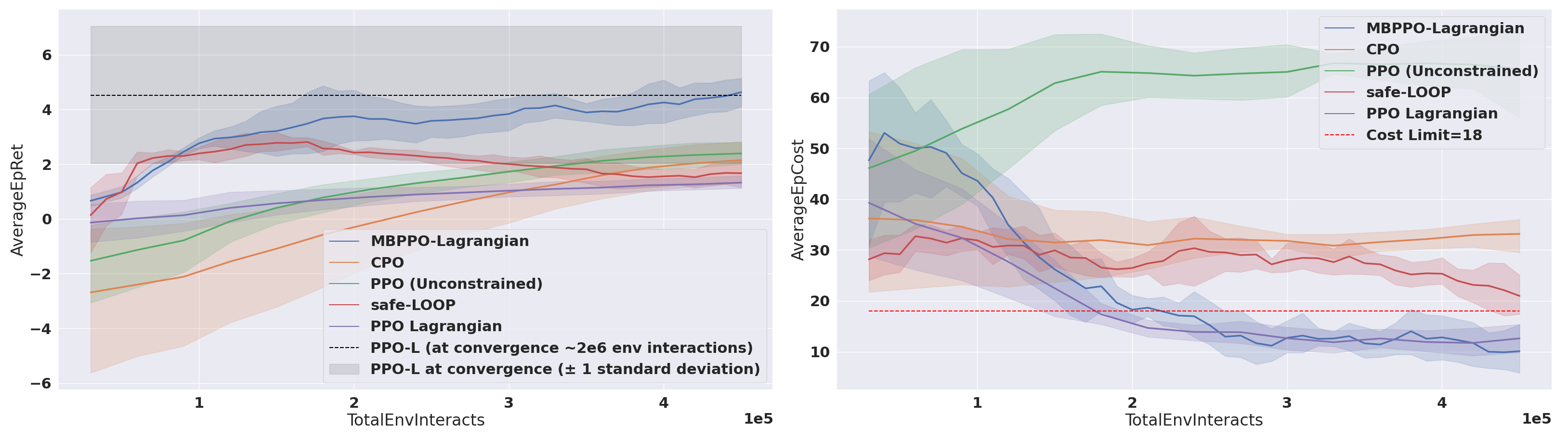}
    \caption{Reward Performance  (Left) and Cost Performance (Right) in PointGoal Environment, where y-axis denotes Average Episode Reward Returns (left) / Cost Returns (right) and x-axis denotes total environment interacts}
    \label{fig:pg2}
\end{figure}

\begin{figure}[H]
    \centering
    \includegraphics[width=0.98\linewidth]{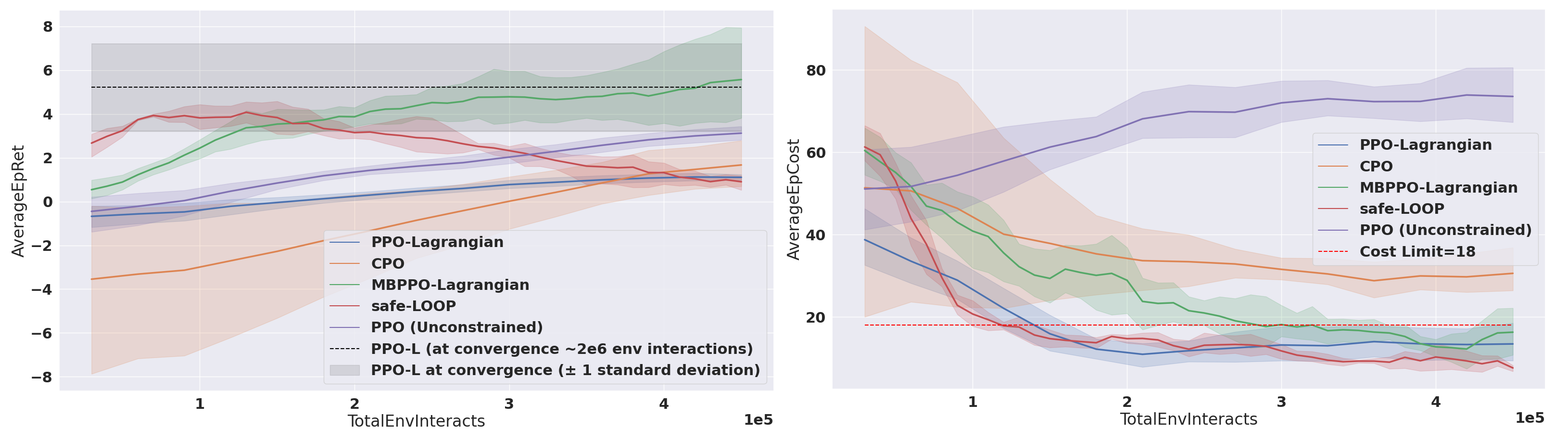}
    \caption{Reward Performance  (Left) and Cost Performance (Right) in CarGoal Environment, where y-axis denotes Average Episode Reward Returns (left) / Cost Returns (right) and x-axis denotes total environment interacts }
    \label{fig:cg2}
\end{figure}

The code for our approach is available here\footnote{\url{https://github.com/akjayant/mbppol}}.  We run each algorithm with 8 random seeds for 450K environment interactions. The hyper-parameter settings and other experimental details are given in Appendix A. 
The performance of our approach and baseline algorithms is shown in Figure \ref{fig:pg2} for PointGoal environment and Figure \ref{fig:cg2} for CarGoal environment.
 The left part of the plots represent Average Episode Return (average of the episode rewards) on y-axis and total environment interactions on the x-axis. The dashed lines represent performance of the PPO-Lagrangian (in black) at convergence which is around 2 million environment interactions. The right part of the plots show Average Episode Cost (average of the episode costs) on y-axis and total environment interactions on the x-axis. The red dashed line in this plot represents cost limit of 18. From the plots we can observe that our approach (MBPPO-Lagrangian) converges to the same level of reward performance as PPO - Lagrangian in just 450K environment interactions and outperforms the  model-based safe-LOOP algorithm \citep{LOOP}. Also our approach gives constraint adhering policies in both the tasks. In addition to above, we run our algorithm for various values of $\beta$ for both \textit{CarGoal} and \textit{RC-Car}\citep{Ahn-2019-117213} in a similar manner as we did for \textit{PointGoal} in Figure \ref{fig:beta} and present it in Appendix D. We also compare our algorithm with safe-LOOP\citep{LOOP} on \textit{RC-Car} environment and present it in Appendix D as well. We found both approaches competitive in \textit{RC-Car} environment, although safe-LOOP \citep{LOOP} exhibits higher variance.
 
\begin{figure}[H]
    \centering
    \includegraphics[width=0.95\linewidth]{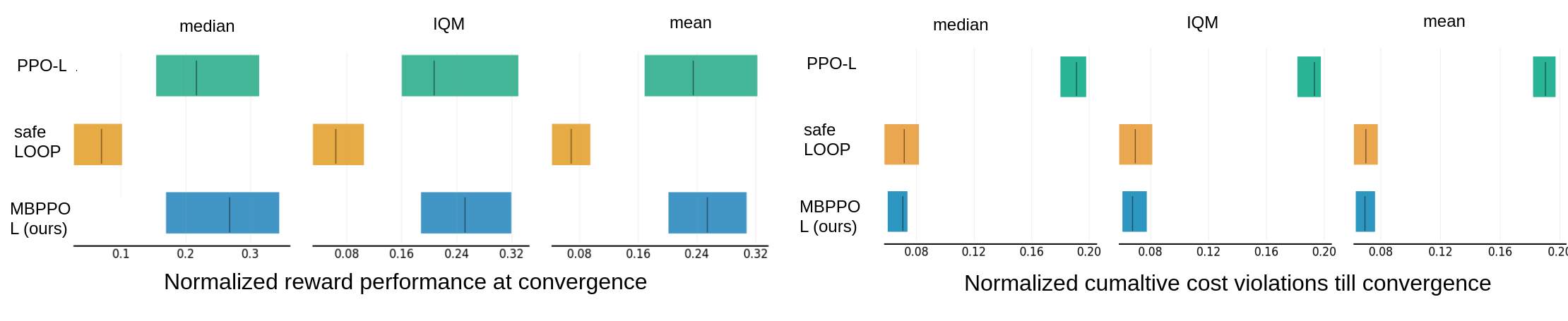}
    \caption{ Normalized Reward Returns at Convergence (left) with median, inter-quartile mean (IQM), mean estimates and Normalized Cumulative Violations (right) with median, inter-quartile mean (IQM), mean estimates. Top rows (in green) represent PPO-Lagrangian, middle rows (in orange) represent safe-LOOP and bottom rows (in blue) represent our approach.}
    \label{fig:error_bars}
\end{figure}

 We also measure cumulative hazard violations that occur  till convergence for MBPPO-L (ours), PPO-Lagrangian \citep{Achiam2019BenchmarkingSE}, safe-LOOP \citep{LOOP} as follows - 
\begin{equation}
\label{countcv}
Cumulative\ Violations = \sum_{Till\ convergence} [\mathbbm{1}(C(s_t,a_t)==1)].
\end{equation}
We use \textit{rliable} library (See Sec 4.3 in \cite{agarwal2021deep} for more details) for plotting $95\%$ confidence intervals for cumulative violations in the right part of Figure \ref{fig:error_bars}  (normalized by cumulative violations in unconstrained PPO in respective tasks) and final policy reward performance in the left part of Figure \ref{fig:error_bars} (normalized by reward return of the final policy in unconstrained PPO in respective tasks) using mean, median and inter-quartile mean as aggregate estimates for our approach,   PPO-Lagrangian and safe-LOOP because these approaches only give constraint satisfying policies at convergence. From figure \ref{fig:error_bars} we can observe that our approach is competitive with regards the model-based safe-LOOP approach \citep{LOOP} in terms of cumulative constraint violations but achieves better reward performance at convergence. 

\section{Conclusions}
We presented a sample efficient approach for Safe Reinforcement Learning and compared our approach with relevant model-free and model-based baselines \citep{achiam2017constrained,LOOP,Achiam2019BenchmarkingSE,schulman2017proximal}.  Note that we chose the above baselines because they are specifically designed to solve a constrained optimization problem of the same structure as (\ref{jr})-(\ref{cop1}). In \cite{LOOP}, approximation of the value function is used to provide long-term reasoning instead of using a truncated horizon. A limitation of this approach lies in the fact that approximations of reward and cost value functions are used. The most challenging part of model-based approaches is to learn the environment model. The first issue is of computational resources and the time overhead that is needed. We provide a comparison of the running time of algorithms and their hardware requirements (see Appendix B). Our algorithm does much better than safe-LOOP \citep{LOOP} in terms of running time. 

One should note that in safe exploration settings, agents do not explore as much as an unconstrained agent would do. This adds to the complexity of model learning in safe exploration problems because constrained exploration leads to limited representation of data points for model learning even with the same sample size. We found this more pronounced in high-dimensional environments like \textit{DoggoGoal1}\citep{Achiam2019BenchmarkingSE} (See Appendix C). Our algorithm depends on Lagrangian-based approach which provides a very simplistic way to construct a safe RL algorithm but suffers from the issue of low reward performance compared to unconstrained approaches (see Appendix A for a reward comparison). Note that we can have a similar model-based approach using TRPO-Lagrangian \citep{Achiam2019BenchmarkingSE} but it involves an extra overhead of approximating the Fisher Information Matrix (FIM) and hence we chose a PPO-based approach. Also TRPO-Lagrangian and PPO-Lagrangian have similar performance on Safety Gym benchmark \citep{Achiam2019BenchmarkingSE}. Increasing reward performance of Lagrangian-based approaches and devising better ways for model-learning in high-dimensional state representations in safe RL settings where exploration is limited, can be looked at in the future. Moreover, it would be interesting to adapt off-policy natural actor-critic algorithms such as in \cite{nac,diddigi} to the setting of constrained MDPs and study their performance both theoretically and empirically.

\begin{ack}
Ashish K. Jayant was supported in his work through an MHRD scholarship from the  Ministry of Education, Government of India; Indian Institute of Science, Bangalore; and  Flipkart Internet Pvt. Ltd., Bangalore. 
S.~Bhatnagar was supported in his work through the J.C.Bose National Fellowship, Science and Engineering Research Board, Government of India; a project 
from the Department of Science and Technology under the ICPS program; a project from DRDO under JATP-CoE; as well as the Robert Bosch Centre for Cyber Physical Systems, Indian Institute of Science, Bangalore.
\end{ack}

\newpage
\bibliographystyle{plainnat}
\bibliography{references}
\section*{Checklist}


\begin{enumerate}

\item For all authors...
\begin{enumerate}
  \item Do the main claims made in the abstract and introduction accurately reflect the paper's contributions and scope?
    \answerYes{}
  \item Did you describe the limitations of your work?
    \answerYes{We describe limitations in Conclusion section}
  \item Did you discuss any potential negative societal impacts of your work?
    \answerNA{}
  \item Have you read the ethics review guidelines and ensured that your paper conforms to them?
    \answerYes{We have read the ethics review guidelines and ensured that our work conforms to them.}
\end{enumerate}

\item If you are including theoretical results...
\begin{enumerate}
  \item Did you state the full set of assumptions of all theoretical results?
    \answerNA{}
        \item Did you include complete proofs of all theoretical results?
    \answerNA{}
\end{enumerate}

\item If you ran experiments...
\begin{enumerate}
  \item Did you include the code, data, and instructions needed to reproduce the main experimental results (either in the supplemental material or as a URL)?
    \answerYes{We provide the github url for our code repository}
  \item Did you specify all the training details (e.g., data splits, hyperparameters, how they were chosen)?
    \answerYes{Hyperparameter details are given in appendix A}
        \item Did you report error bars (e.g., with respect to the random seed after running experiments multiple times)?
    \answerYes{We report plots with 95\% confidence intervals and use rliable \cite{agarwal2021deep} library for robust aggregate estimates}
        \item Did you include the total amount of compute and the type of resources used (e.g., type of GPUs, internal cluster, or cloud provider)?
    \answerYes{We share details of hardware requirement in appendix B}
\end{enumerate}

\item If you are using existing assets (e.g., code, data, models) or curating/releasing new assets...
\begin{enumerate}
  \item If your work uses existing assets, did you cite the creators?
    \answerYes{We cite the authors of Safety Gym benchmark and their github code repository.}
  \item Did you mention the license of the assets?
    \answerYes{Their MIT License which allows  without restriction, including without limitation the rights
to use, copy, modify, merge, publish provided that license notice is included in our code repository}
  \item Did you include any new assets either in the supplemental material or as a URL?
    \answerNA{}
  \item Did you discuss whether and how consent was obtained from people whose data you're using/curating?
    \answerNA{}
  \item Did you discuss whether the data you are using/curating contains personally identifiable information or offensive content?
    \answerNA{}
\end{enumerate}

\item If you used crowdsourcing or conducted research with human subjects...
\begin{enumerate}
  \item Did you include the full text of instructions given to participants and screenshots, if applicable?
    \answerNA{}
  \item Did you describe any potential participant risks, with links to Institutional Review Board (IRB) approvals, if applicable?
    \answerNA{}
  \item Did you include the estimated hourly wage paid to participants and the total amount spent on participant compensation?
    \answerNA{}
\end{enumerate}

\end{enumerate}


\appendix
\section{Hyper-parameters and finer experimental details}

The hyper-parameters used for our algorithm are shown in Table \ref{hyp}. 
 \begin{table}[h]
  \caption{Hyper-parameters used for our algorithm MBPPO-Lagrangian}
  \label{hyp}
  \centering
  \begin{tabular}{lll}
    \toprule
    Hyperparameter     & Value/Description \\    
    \midrule
No. of models in ensemble $(n)$ & 8\\

Hidden layers in single ensemble NN & 4\\

(Hidden layers in single ensemble NN : no. of nodes) & (200:200:200:200)\\
Hidden layers in Actor NN & 2\\

(Hidden layers in Actor NN : no. of nodes) & (64:64)\\

Hidden layers in Critic NN  & 2\\

(Hidden layers in Critic NN : no. of nodes) & (64:64)\\

Gradient Descent Algorithm & ADAM\\

Actor Learning Rate (for $\theta$ update) & 3e-4\\

Critic Learning Rate (for $\psi_r,\psi_c$ update) & 1e-3\\

Lagarange Multiplier Learning Rate (for $\lambda$ update) & 5e-2\\

Initial Lagrange Multiplier Value $(\lambda_0)$ & 1\\

Cost limit $(d)$ & 18\\

Activation Function & $tanh$\\

Discount Factor $(\gamma)$ & 0.99\\

GAE parameter ($\bar{\lambda}$) & 0.95\\

Horizon  ($H$) in step 7 of Algorithm 1 & 80\\

Validation Dataset/Train Dataset & 10\%/90\%\\

PR Threshold & 66\%\\

Mix of real data and imaginary data for first pass & 5\%95\%\\

$\beta$ used in (29)  & 0.02\\
    \bottomrule
  \end{tabular}
\end{table}
Each episode runs for 750 steps as opposed to 1000 steps in the original version of Safety Gym, hence we chose cost threshold of 18 which is $\sim75\%$ of 25 used in the official Safety Gym benchmark paper. In Goal-based Safety gym environments the aim is to reach the goal position (in green in Figure 1) with as few collisions as possible with hazards. If robot (in red in Figure 1) accesses 'hazard' positions (in blue in Figure 1), the agent incurs a cost = 1. The 'Point' robot has steering and throttle as action space while 'Car' robot has differential control. We use  Performance Ratio (PR) threshold of 66\%.  The logic behind this number is that our agent should perform better in more than $50\%$ of the models in the ensemble. In our case we train an ensemble of 8 models out of which we use the best 6 models with minimum validation losses to calculate PR. We want our agent to perform better in at least 4 out of the 6 models that gives us the value as $66\%$. As mentioned in the text we increase the number of hazards in PointGoal1 and CarGoal1 environments from 10 to 15 so as to increase their difficulty level. 
\begin{figure}[h]
    \centering
    \includegraphics[width=0.45\linewidth]{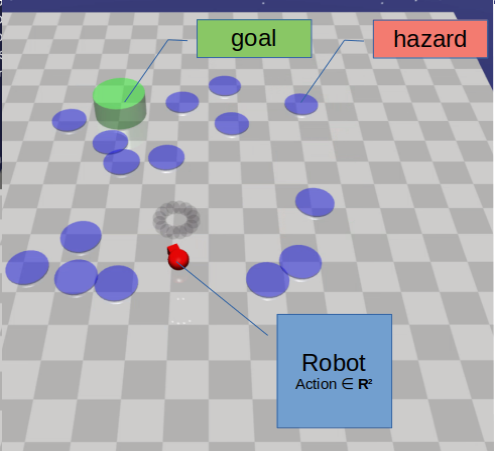}
    \caption{Safety Gym with labeled elements}
    \label{fig:sg}
\end{figure}
We run on 8 random seeds which is greater than 4 used in benchmarking previous unconstrained model-based approaches \footnote{\url{https://github.com/WilsonWangTHU/mbbl}}. We use the official implementation of baselines from safe-LOOP\footnote{\url{https://github.com/hari-sikchi/LOOP}},  CPO\footnote{\url{https://github.com/jachiam/cpo}} and PPO-Lagrangian \footnote{\url{https://github.com/openai/safety-starter-agents}}, respectively. In all baselines we use the same size of the neural networks as mentioned in Table \ref{hyp} and the same number of models in ensemble for model learning for safe-LOOP. In safe-LOOP, we do not change the originally used planning horizon of $H=8$ and cost threshold of 0 in the model optimization part. During first pass of Algorithm 1 (Line 5-17), we use $5\%$ of real environment interaction data and $95\%$ of imaginary rollout data, this is done to make use of real environment data that we collected to train model dynamics. After first pass, policy gets updated and after that we use purely imaginary data since this is an on-policy algorithm until PR goes below $66\%$.

The values used for normalization in plot for Figure 4 are the following: Reward performance of PPO-unconstrained = 20 and Cumulative constraint violations of PPO-unconstrained = 2e5.

\section{Hardware requirements and Running time}
Minimum 4 GB GPU space is required for running both the model based approaches. 
\begin{table}[h]
  \caption{Running time for 450k steps (in seconds)}
  \label{run}
  \centering
  \begin{tabular}{lll}
    \toprule
    Algorithm     & Running time (in s) \\    
    \midrule
PPO-Lagrangian & 187.95 $\pm$ 7.56  \\
CPO & 266.25 $\pm$ 6.46  \\
MBPPO-Lagrangian  & 21420.91 $\pm$ 554.449\\
safe-LOOP & 183156.33 $\pm$ 19083.43 \\
\end{tabular}
\end{table}
We did not use multiprocessing in model-based approaches for data-collection from simulation. However, for model-free approaches we collected simulation data using 8 CPUs at a time. Running time for 450k steps for all baselines is given in Table 2.

Note that for convergence PPO-Lagrangian required 849 $\pm$ 12.48 s and  CPO required 1162 $\pm$ 8.87 s. The running time of our model-based approach can be improved by using parallel workers for data collection but still major part of training is spent in model learning.

\section{Comparison of model learning validation loss (Unconstrained vs Safe RL) and performance in High-dimensional task}
\begin{figure}[h]
    \centering
    \includegraphics[width=0.75\linewidth]{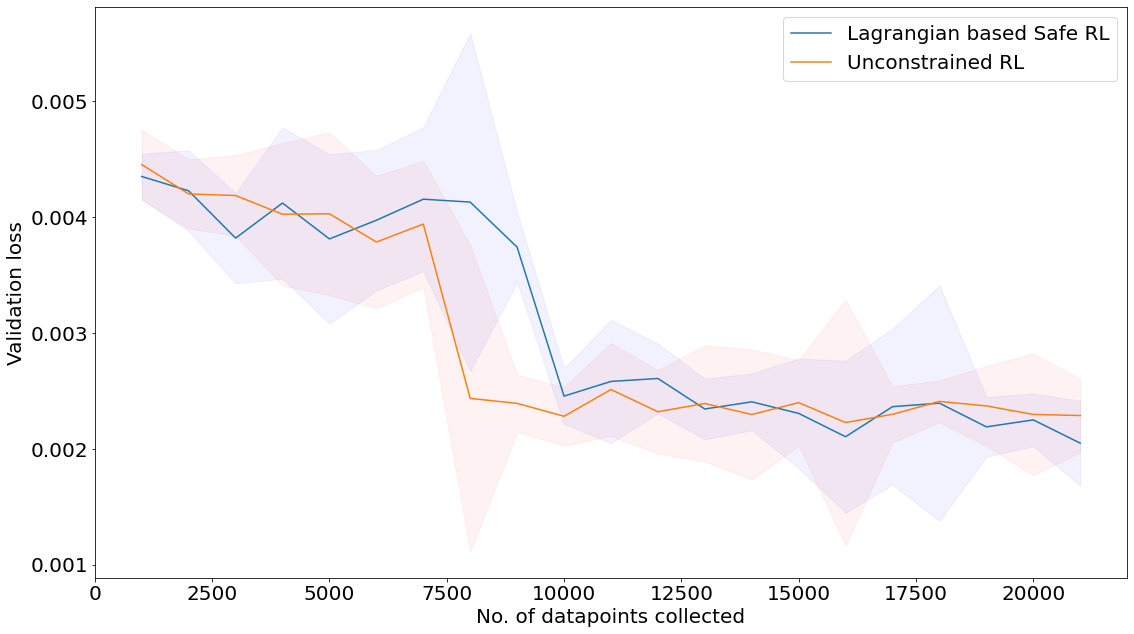}
    \caption{Quicker convergence of Validation loss in unconstrained setting as compared to constrained setting}
    \label{fig:uc2}
\end{figure}
We compare how model learning validation loss varies in Safe RL setting as opposed to unconstrained RL one. We plot validation loss vs no. of datapoints collected as our model-based algorithm progress (See Figure \ref{fig:uc2}. We observe that validation error converges quickly in unconstrained setting, which is reflective of better exploration in unconstrained setting. This plot is generated using 8 different initialization of neural network used for model learning. We do this for our modified PointGoal1 environment.

\textit{DoggoGoal1} is a challenging environment where  a quadruped robot has to reach a goal position without colliding with obstacles. This task has higher dimensional action space (12-dimensional) than Point  and Car robots  which just needs steering and throttle in Safety Gym. On this task, even model-free Lagrangian baselines suffer and get quite low reward performance. Further, both safe-LOOP and MBPPO-Lagrangian fails to match model-free baseline in terms of reward performance in DoggoGoal environment. It signifies the additional challenge in model learning in case of high-dimensional environments. 
\begin{table}[h]
  \caption{Reward Performance at convergence in DoggoGoal}
  \label{rewdoggo}
  \centering
  \begin{tabular}{lll}
    \toprule
    Algorithm     &  Episodic reward performance of final policy\\    
    \midrule
PPO  &  21.3 $\pm$ 1.23  \\
PPO-Lagrangian &  1.6275 $\pm$ 0.46  \\
safe-LOOP & -0.14 $\pm$ 0.05 \\
MBPPO-Lagrangian  & -0.69 $\pm$ 0.06 \\
\end{tabular}
\end{table}
\section{Effect of $\beta\ $ in  CarGoal and RC-Car environment and Model-based baselines results on RC-Car}
In RC Car\footnote{\url{https://github.com/r-pad/aa_simulation}} environment,
 a car has to rotate within a circle with target velocity for earning rewards. If car goes out of circle, it incurs cost.
\begin{figure}[h]
    \centering
    \includegraphics[width=0.30\linewidth]{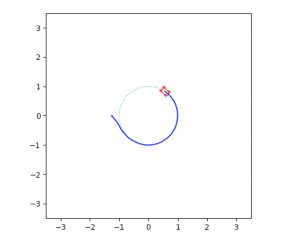}
    \caption{Car in (red) has to drive along the circle of fixed radius with some target velocity}
    \label{fig:rc}
\end{figure}

 \begin{figure}[h]
    \centering
    \includegraphics[width=0.80\linewidth]{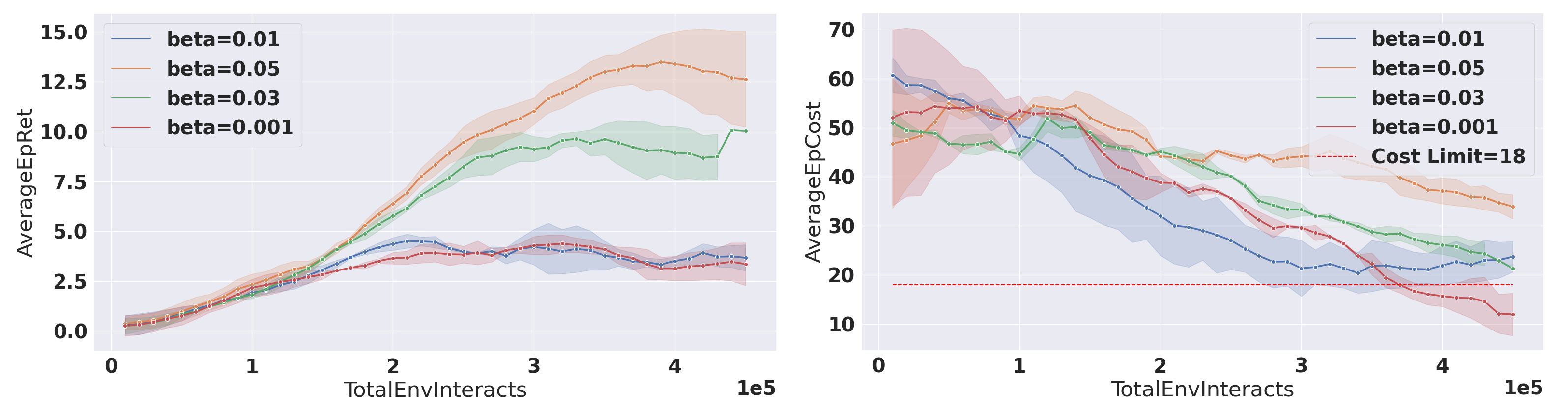}
    \caption{Effect of $\beta$ in CarGoal Environment on Reward returns (Left) and Cost returns (Right)}
    \label{fig:betacar}
\end{figure}
\begin{figure}[H]
    \centering
    \includegraphics[width=0.80\linewidth]{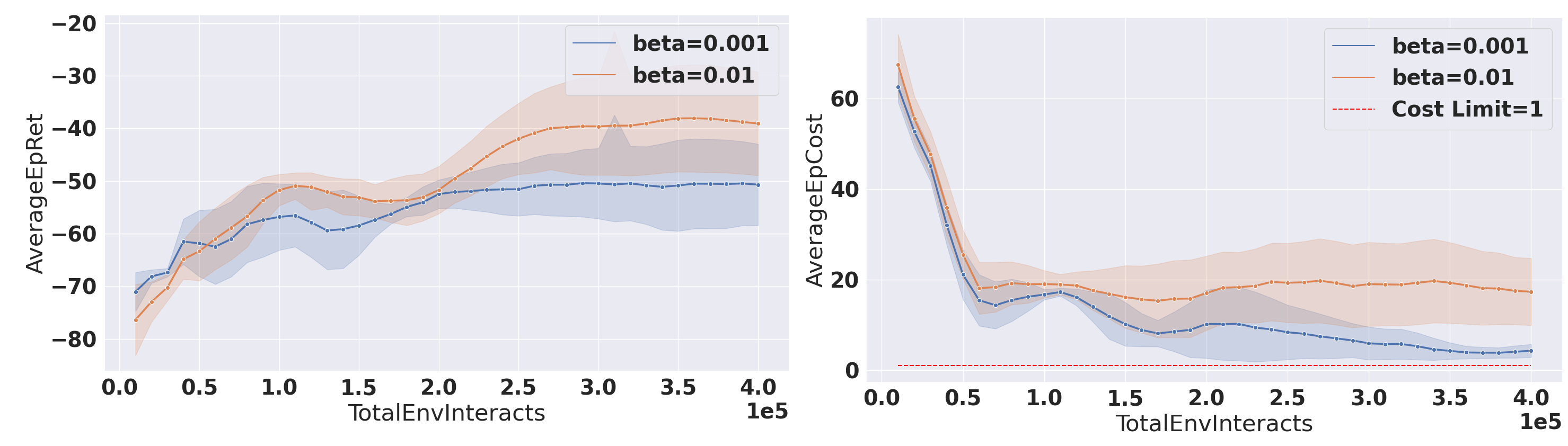}
    \caption{Effect of $\beta$ in RC-Car Environment on Reward returns (Left) and Cost returns (Right)}
    \label{fig:betarc}
\end{figure}
\begin{figure}[H]
    \centering
    \includegraphics[width=0.85\linewidth]{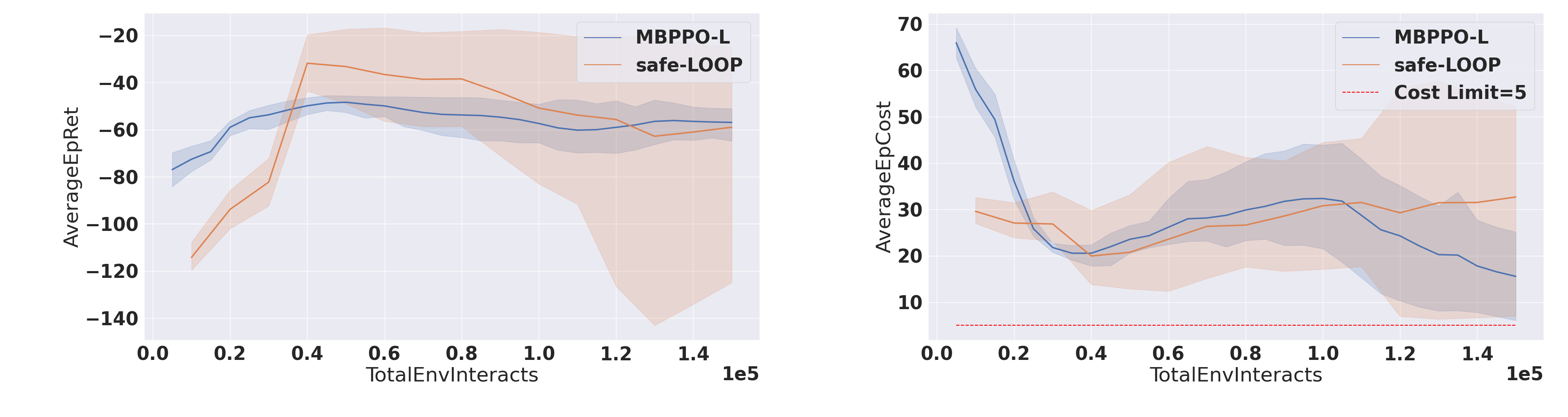}
    \caption{Reward returns (Left) and Cost returns (Right) in RC-Car environment}
    \label{fig:rc_mb}
\end{figure}

  We observe a similar trend as observed in Figure 1 of the paper, i.e., lower the value of $\beta$, lower are the reward returns as the agent explores pessimistically. We also ran MBPPO-L(our algorithm) and safe-LOOP \citep{LOOP} on the \textit{RC-Car} environment for 150k steps (with 5 random seeds) only as safe-LOOP has a very high running time. We choose cost limit = 5. Results are presented in Figure \ref{fig:rc_mb}. We can observe that both approaches have  similar performance but safe-LOOP exhibits higher variance than our approach.
\end{document}